\documentclass[letterpaper, 10 pt, conference]{ieeeconf}  % Comment this line out if you need a4paper

\IEEEoverridecommandlockouts                              % This command is only needed if 
\usepackage{graphics} % for pdf, bitmapped graphics files
\usepackage{epsfig} % for postscript graphics files
\usepackage{times} % assumes new font selection scheme installed
\usepackage{amsmath} % assumes amsmath package installed
\usepackage{amssymb}  % assumes amsmath package installed
\usepackage{cite}
\usepackage{mathtools}
\usepackage{multirow}
\usepackage{booktabs}
\usepackage{caption}
\usepackage{subcaption}
\usepackage{hyperref}
\hypersetup{
    colorlinks=true,
    urlcolor=blue
}
\usepackage{xcolor}
\usepackage{cleveref}
\renewcommand{\bf}[1]{\textbf{#1}}      % shortcut for bold font
\renewcommand{\it}[1]{\textit{#1}}      % shortcut for italic font
\newcommand{\R}{\mathbb{R}}           % shortcut for the set of real numbers
\newcommand{\commentout}[1]{}

\newcommand{\cvec}[1]{\boldsymbol{\mathrm{#1}}}

\title{\LARGE \bf
UniFField: A Generalizable Unified Neural Feature Field for\\ Visual, Semantic, and Spatial Uncertainties in Any Scene
}

\author{Christian Maurer$^{*1}$, Snehal Jauhri$^{*1}$, Sophie Lueth$^{1}$, Georgia Chalvatzaki$^{1,2,3}$\\
$^{*}$ indicates equal contribution\\
$^{1}$TU Darmstadt $^{2}$Hessian.AI $^{3}$Robotics Institute Germany\\
\thanks{- All authors are with the Computer Science Department, Technische Universit\"{a}t Darmstadt, Germany:
        {\texttt{\{christian.maurer, snehal.jauhri, sophie.lueth\}@tu-darmstadt.de},
\texttt{georgia.chalvatzaki@tu-darmstadt.de}
}}%
\thanks{- Research funded by EU Horizon program under grant no. 101120823, project MANiBOT. Support and HPC resources provided by Erlangen National High Performance Computing Center (NHR) of Friedrich-Alexander-Universität Erlangen-Nürnberg (FAU), funded by federal and Bavarian authorities and the German Research Foundation (DFG) – 440719683.}% <-this % stops a space
}

\let\oldtwocolumn\twocolumn
\renewcommand\twocolumn[1][]{%
    \oldtwocolumn[{#1}{
    \vspace{-1.2cm}
    \begin{center}
        \includegraphics[width=\textwidth]{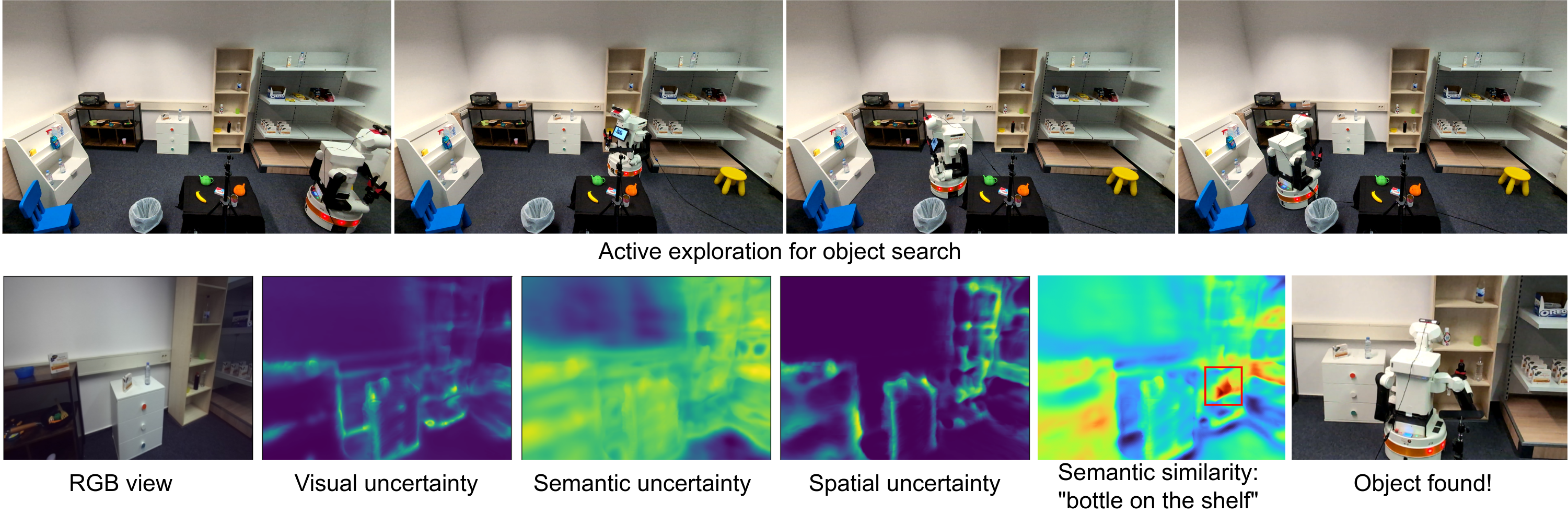}
        \captionof{figure}{\textbf{Example object search task using UniFField}. The robot explores the scene and incrementally builds the volumetric UniFField representation. UniFField enables uncertainty-aware feature prediction in each modality, thus enabling weighted similarity search based on a language query to find the target object. Project website: \href{https://sites.google.com/view/uniffield}{https://sites.google.com/view/uniffield}
        }
        \label{fig:cover}
    \end{center}
    }]
}

\begin{document}

\maketitle
\thispagestyle{empty}
\pagestyle{empty}

%%%%%%%%%%%%%%%%%%%%%%%%%%%%%%%%%%%%%%%%%%%%%%%%%%%%%%%%%%%%%%%%%%%%%%%%%%%%%%%%
\begin{abstract}
Comprehensive visual, geometric and semantic understanding of a 3D scene is crucial for successful execution of robotic tasks, especially in unstructured and complex environments. Additionally, to make robust decisions it is necessary for the robot to evaluate the reliability of perceived information.
% Robot perception requires effective 3D scene understanding from visual, geometric, and semantic perspectives, particularly in unstructured environments where robots must quickly process information and make decisions. In addition, the robot must also be able to infer the reliability of perceived information to minimize risks and ensure robust performance. 
While recent advances in 3D neural feature fields have enabled robots to leverage features from pretrained foundation models for tasks such as language-guided manipulation and navigation, existing methods suffer from two critical limitations: (i) they are typically scene-specific, and (ii) they lack the ability to model uncertainty in their predictions. We present \bf{UniFField}, a unified uncertainty-aware neural feature field that combines visual, semantic, and geometric features in a single generalizable representation while also predicting uncertainty in each modality. 
Our approach, which can be applied zero shot to any new environment, incrementally integrates RGB-D images into our voxel-based feature representation as the robot explores the scene, simultaneously updating uncertainty estimation. We evaluate our uncertainty estimations to accurately describe the model prediction errors in scene reconstruction and semantic feature prediction. Furthermore, we successfully leverage our feature predictions and their respective uncertainty for an active object search task using a mobile manipulator robot, demonstrating the capability for robust decision-making.
\end{abstract}

%%%%%%%%%%%%%%%%%%%%%%%%%%%%%%%%%%%%%%%%%%%%%%%%%%%%%%%%%%%%%%%%%%%%%%%%%%%%%%%%

%\section{INTRODUCTION}
\section{INTRODUCTION}

Generalist robots that can adapt to any environment, whether a cluttered living room or a busy kitchen, represent the next frontier in robotics. Such robots require effective 3D perception to quickly understand scenes, make decisions, and act. To this end, there has been significant interest in building 3D neural representations for robots by distilling features from 2D vision encoders and foundation models into 3D~\cite{shen2023F3RM}. This enables robots to leverage prior, pretrained information for tasks such as language-guided manipulation and navigation~\cite{lerftogo2023, wang2024d3fields, chen2025splatnav, chen2025g3flow}. However, most 3D neural feature fields are scene-specific, i.e., they are trained using a fixed set of 2D images and their respective features captured for a single scene. Moreover, a significant drawback of these techniques, which use NeRF or Gaussian Splatting representations, is the inability to incrementally add observations as the robot explores the scene, which is crucial for robots that need to operate in unknown or quickly changing environments.

Recently, attempts have been made to learn general-purpose feature fields for robots that can be pretrained on multiple scenes and then be applied zero shot to \textit{any} scene~\cite{qiu-hu-song-2024-geff}. Moreover, some recent works have also focused on learning incremental neural representations that can aggregate information over time~\cite{shafiullah2022clipfields, yu2024language}. However, a key missing piece for such 3D feature representations is the ability to model the reliability or uncertainty of perceived scene features. Such uncertainty can be crucial for continuous robot perception in real-world scenarios where observations can be noisy, partial, and only parts of objects can be briefly seen. Moreover, most 2D vision feature encoders such as CLIP~\cite{radford2021learning} or DINO~\cite{oquab2023dinov2} can still be very noisy in their predictions, especially in partially observable settings. Therefore, it is crucial to have a 3D feature representation that can also model uncertainty in the features, which can be used for downstream tasks such as active perception and exploration. 
%previously: Especially for active exploration, uncertainties can exist due to part of the scene being unexplored, due to the model's lack of prior knowledge from the data it was trained on (epistemic uncertainty), or due to inherent difficulties in predicting the semantic or geometric features (aleatoric uncertainty).
Especially for active exploration, uncertainties can exist when parts of the scene remain unexplored, from model uncertainty due to limited knowledge or insufficient training data (epistemic uncertainty), or from inherent noise in the observations used for learning (aleatoric uncertainty).

This work introduces UniFField, a unified uncertainty-aware neural feature field for 3D scene understanding from multi-view RGB-D data. Our 3D feature field combines visual, semantic, and geometric features in one representation while also predicting uncertainty in each modality. %Our representation is aggregating, i.e., it allows incremental updates, and is generalizable to any scene zero shot. We validate the ability of our representation in predicting accurate uncertainties in the scene that appropriately co-relate with prediction errors of the scene geometry or semantics.We also demonstrate the effectiveness of uncertainty prediction for active object search using a mobile manipulator robot.

Our main contributions are as follows,
\begin{itemize}
    \item We propose a generalizable unified neural feature field, UniFField, that provides a prior for visual, semantic, and geometric feature predictions. Semantic information is integrated by distilling 2D vision-language features into the 3D representation.
    \item We use UniFField to model uncertainty in each modality, enabling robust decision making in partially observable settings. Our predicted uncertainties accurately describe the prediction errors of the model.
    \item Our representation lifts features from 2D to 3D while also being aggregating, i.e., it allows incremental updates, ideal for robots continuously exploring scenes.
    \item We devise a simple but effective approach to using the uncertainty-aware UniFField for an active object search task using a mobile manipulator robot.
\end{itemize}

% \textbf{Notes:}
% \begin{itemize}
%     \item Uncertainty-awareness and robustness against sparse or
%     ambiguous observations.
%     \item Demonstrating the effectiveness of the predicted uncertainties in describing the model’s prediction errors.
%     \item Integration of uncertainty quantification into the perception systems of future robots, enabling more informed and robust decision-making.
% \end{itemize}

%\section{RELATED WORK}
\section{RELATED WORK}

\textbf{Geometric Reconstruction}. 
Multi-view geometric scene reconstruction methods can be divided into (i)~Depth-based methods \cite{long2021multiview, watson2021temporal} that estimate per-view depth maps, merge them via volumetric fusion~\cite{curless1996volumetric, izadi2011kinectfusion}, and ensure consistent surface representation~\cite{wang2021multiview, wang2024learningbased}; and (ii)~Volumetric methods~\cite{stier2023finerecon, li2025hybrid} that operate on dense 3D grids for occupancy or signed distance prediction. High computation time remains a challenge for both approaches, with recent work focusing on high-quality real-time reconstruction both for depth-based~\cite{sayed2022simplerecon, sayed2024doubletake} and volumetric methods~\cite{li2025hybrid}.
Recent works have combined both concepts~\cite{stier2023finerecon, feng2023cvrecon} to overcome their downsides, namely low prediction quality in areas of few feature points~\cite{long2022sparsenues}, floating artifacts due to lack of global consistency~\cite{sayed2022simplerecon} in depth-based methods, and inadequate modeling of view-dependent information in volumetric methods~\cite{sayed2022simplerecon}. 

\textbf{Generalizable Priors}.
Incorporating learned priors into neural fields can improve prediction quality of geometric reconstruction or other modalities like semantics. While geometric priors significantly alleviate reconstruction challenges~\cite{azinovic2022neural, guo2022manhattan, lee2023fastsurf}, they still require training a separate model from scratch for every scene. In contrast, generalizable scene priors~\cite{liu2023one2345, liu2023one2345pp} learned from large-scale datasets generalize better across unseen scenes~\cite{fu2023scenepriors}, allowing for fast and robust reconstruction even with limited input views~\cite{long2022sparsenues}. Generalizable Priors can also be used for understanding the semantics of previously unseen 3D scenes~\cite{qiu-hu-song-2024-geff, ze2023gnfactor,chen2025g3flow, wang2024govnesf, chou2024gsnerf}.

\textbf{Semantic Scene Understanding}.
Feature Fields extend Neural Fields, which combine the encoding of image and spatial representations of a scene, with additional modalities like semantic information~\cite{oquab2023dinov2, radford2021learning}, e.g. with language~\cite{roessle2022dense, lerftogo2023, wang2024govnesf}. The combination of geometric and semantic features can improve performance, leveraging each others' consistency~\cite{chou2024gsnerf}.
Featurenerf~\cite{ye2023featurenerf} utilizes this property to transform part segmentation labels and key-points to different views. Other approaches~\cite{jatavallabhula2023conceptfusion, peng2023openscene, wang2024govnesf} enable open-vocabulary and zero-shot spatial reasoning for tasks like 3D semantic segmentation and 3D object search. 

\textbf{Uncertainty Quantification for Neural Radiance Fields}. % add an introductory sentence that uncertainty is not modeled?
Estimating uncertainty of neural representations can be used for decision making in active perception pipelines~\cite{jauhri2024active}. While the predicted uncertainty can be modeled directly as a Gaussian distribution over outputs~\cite{roessle2022dense}, we aim to learn a prior over the uncertainty in the training data. Distractor-free NeRFs separate scenes into static and dynamic components with, e.g., the help of semantic features~\cite{yang2023emernerf, ren2024nerfonthego, wang2024d3fields}. Similarly, we also integrate additional uncertainty indicators obtained from the input data to improve overall uncertainty quantification. For radiance fields, variational Bayesian methods can be used to model distribution~\cite{shen2021stochastic, shen2022conditional}. We leverage approximate Bayesian methods like Dropout~\cite{lakshminarayanan2017simple} and Ensembles~\cite{gal2016dropout, sunderhauf2023density}, that have been adapted for NeRFs~\cite{murez2020atlas}. Bayes' Rays~\cite{goli2023bayesrays} estimates epistemic uncertainty post training of NeRFs by learning a volumetric field of allowed spatial perturbations that do not degrade reconstruction quality.
%EvidMTL~\cite{menon2025iros} explicitly models uncertainty for depth and semantics using evidential multi-task learning and performs uncertainty-aware semantic TSDF mapping from monocular RGB by evidentially integrating observations and jointly updating semantic and TSDF posteriors in 3D.
EvidMTL~\cite{menon2025iros} jointly predicts semantics and depth with explicit aleatoric and epistemic uncertainties, enabling uncertainty-aware 3D semantic monocular TSDF mapping.

We choose a hybrid geometric approach similar to~\cite{stier2023finerecon} and enrich our accumulated volumetric features~\cite{murez2020atlas} with depth maps. We learn generalizable semantic priors similar to GeFF~\cite{qiu-hu-song-2024-geff} and leverage a voxel-based feature representation~\cite{ze2023gnfactor}. Despite recent advances in generalizable neural radiance fields, their ability to quantify uncertainty remains limited. In this work, we bridge this gap with UniFField.

%Question: should we write this?
% Notably, to the best of our knowledge does no prior work combine geometric and semantic feature fields with uncertainty

% \textbf{Active Perception}.

% In contrast to geometric reconstruction from multi-view images, the expressiveness of Neural Fields is not limited by fixed image resolutions but instead allows for querying the surface representation at arbitrary points in space.

% \textbf{TODO: CITE all papers including recent CoRL ones}
% \begin{itemize}
%     \item DeepSDFs \& NERFs
%     \item Featrue Nerfs / Gen feature fields
%     \item Language features in scene representations \cite{peng2023openscene}
%     \item Generalizable vs. Scene specific
%     \item Aggregating neural representations
% \end{itemize}

%\section{VISUAL-FF}
\section{UniFField}
% \textbf{Notes:}
% \begin{itemize}
%     \item 2D to 3D Atlas-style representation
%     \item Fine-Recon-style MLPs
%     \item Distillation
%     \item Supervising the feature field
%     \item Phase-based training (Pre-training vs Inference)
%     \item Aggregation at training time vs inference time 
%     \item How we calculate uncertainty: Indicators vs Predicted Uncertainty
% \end{itemize}

%An overview of our proposed scene representation VISUAL-FF is depicted in \Cref{fig:overview}.
% We propose UniFField, a unified scene representation as depicted in~\Cref{fig:overview}.
% In the following, we define the problem and explain how it is constructed and supervised.

% \subsection{Problem definition}

\begin{figure}
    \centering
    \includegraphics[width=0.485\textwidth]{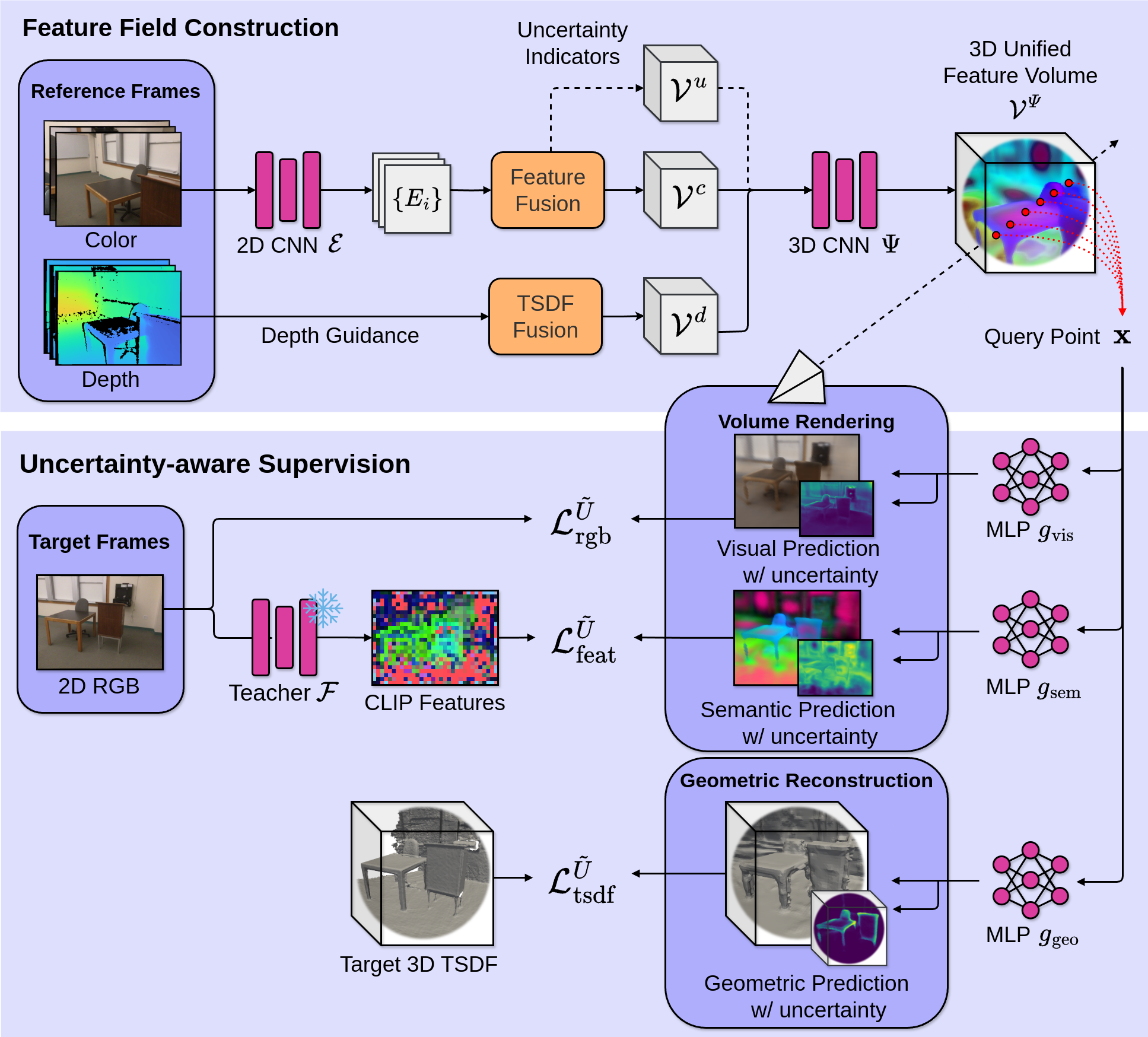}
    \caption[Overview of UniFField]{\textbf{Overview of UniFField}. 
    Given a sequence of \mbox{RGB-D} reference frames of a scene, we combine image features $\mathcal{V}^c$, an initial TSDF volume $\mathcal{V}^d$, and uncertainty indicators $\mathcal{V}^u$ to construct a unified feature volume $\mathcal{V}^{\Psi}$. 
    We employ knowledge distillation of a teacher model~$\mathcal{F}$, novel view synthesis, and geometric reconstruction as pre-training objectives to build the generalizable model. At test time, the model generates visual, spatial, and semantic scene properties, along with their associated uncertainty.
    }
    \vspace{-0.8em}
    \label{fig:overview}
\end{figure}

We address the problem of creating an uncertainty-aware scene representation that can serve as a foundational component for robotic perception, without per-scene optimization.
Given $N$ posed \mbox{RGB-D} frames $\mathcal{D} = \{ ( I_i, D_i, P_i, K_i ) \}_{i=1}^N$, with color images $I_i \in \R^{H \times W \times 3}$, depth maps $D_i \in \R^{H \times W}$, camera poses $K_i \in \R^{3 \times 3}$, and camera intrinsics $P_i \in \mathrm{SE}(3)$, we design a unified feature field
\begin{equation}
    \Phi(\cvec{x}; \mathcal{D}) : \R^3 \mapsto \R^{C_{\Psi}}
\end{equation}
conditioned on $\mathcal{D}$. We map every point $\cvec{x} \in \R^3$ to a unified feature of dimension $C_{\Psi}$ that describes the visual, spatial, and semantic properties of the scene, as well as the corresponding uncertainty. The field is implicit, i.e., queryable at any arbitrary 3D location, allowing for flexible extraction of information at any spatial point. Finally, the field is additive, i.e., allows incremental updates as new RGB-D frames $\mathcal{D}$ are observed in the scene~(\Cref{fig:overview}).%Finally, the model should allow for incremental updates when new observations are added.

\subsection{Constructing a Unified Feature Field}

%We propose VISUAL-FF, as depicted in \Cref{fig:overview}, that
We build a feature volume $\mathcal{V}^{\Psi} \in \R^{V_{\mathrm{x}} \times V_{\mathrm{y}} \times V_{\mathrm{z}} \times C_{\Psi}}$ that structures the scene as a 3D voxel grid with spatial dimensions $V_{\mathrm{x}}, V_{\mathrm{y}}, V_{\mathrm{z}}$. First, we extract dense image features $E_i=\mathcal{E}(I_i) \in \R^{H \times W \times C_{\mathcal{E}}}$ from each RGB image using a 2D CNN encoder $\mathcal{E}$ \cite{murez2020atlas}. Every pixel’s feature is then back-projected along its viewing ray, assigning that feature to all voxels intersected by the ray. This creates an image feature volume~$\mathcal{V}^c$, averaged over all accumulated observations~$\mathcal{D}$, as in~\cite{murez2020atlas}. To further inform the network of the precise spatial locations where features should be assigned, we use depth guidance~\cite{stier2023finerecon} from the input depth channel. We apply the standard TSDF fusion~\cite{curless1996volumetric} algorithm on the depth channel to acquire an initial TSDF (Truncated Signed Distance Function) volume~$\mathcal{V}^d$ of the scene.% In our evaluations, we assess the effect of depth guidance by ablating it.

% To guide the back-projection and enhance the reconstruction accuracy, we acquire an initial TSDF volume from depth maps.
Additionally, to guide the downstream uncertainty predictions of the network, we add two more input signals: voxel-wise feature count and feature variance. The feature count is the number of observations accumulated in each voxel in the feature volume, while feature variance is the variance of those features across observations. Both signals essentially provide metadata from the input feature fusion process, serving as indicators of uncertainty over the volume:~$\mathcal{V}^u$.

We concatenate all volumes into $\mathcal{V} = [ \mathcal{V}^c, \mathcal{V}^d, \mathcal{V}^u ]$ and refine the combined features using a 3D CNN $\Psi$ \cite{stier2023finerecon} to produce the final unified feature volume $\mathcal{V}^{\Psi} = \Psi(\mathcal{V})$.
We apply trilinear interpolation on the feature volume to create the feature field $\Phi(\cvec{x}; \mathcal{D}) \coloneqq \mathrm{Trilinear}(\mathcal{V}^{\Psi}, x)$, allowing us to query unified features at any continuous 3D location $\cvec{x}$.

%For any spatial 3D point $\cvec{x}$, we trilinearly interpolate the feature volume to obtain a unified feature $\Phi(\cvec{x}; \mathcal{D}) \coloneqq \mathrm{Trilinear}(\mathcal{V}^{\Psi}, x)$.

\subsection{Decoding the Unified Feature Field}

To decode the feature field, we construct three decoding networks on top of the feature field, with their outputs modeled as the mean and variance of Gaussian distributions. Specifically, we predict
\begin{equation}
    \begin{aligned}
        \big( c(\cvec{x}),\; u_c(\cvec{x}) \big) & \coloneqq g_{\mathrm{vis}}(\Phi(\cvec{x}; \mathcal{D})), \\ 
        %\big( \bar{c}(\cvec{x}),\; u_c(\cvec{x}) \big) & \coloneqq g_{\mathrm{vis}}(\Phi(\cvec{x}; \mathcal{D})), \\ 
        \big( f(\cvec{x}),\; u_f(\cvec{x}) \big) & \coloneqq g_{\mathrm{sem}}(\Phi(\cvec{x}; \mathcal{D})), \\
        %\big( \bar{f}(\cvec{x}),\; u_f(\cvec{x}) \big) & \coloneqq g_{\mathrm{sem}}(\Phi(\cvec{x}; \mathcal{D})), \\ 
        \big( s(\cvec{x}),\; u_s(\cvec{x}) \big) & \coloneqq g_{\mathrm{geo}}(\Phi(\cvec{x}; \mathcal{D})), \\ 
        %\big( \bar{s}(\cvec{x}),\; u_s(\cvec{x}) \big) & \coloneqq g_{\mathrm{geo}}(\Phi(\cvec{x}; \mathcal{D})), \\ 
    \end{aligned}
\end{equation}
where $g_{\mathrm{vis}}$, $g_{\mathrm{sem}}$, and $g_{\mathrm{geo}}$ are visual, semantic, and geometric networks implemented as MLPs with two heads. They map a unified feature at 3D point $\cvec{x}$ to the mean RGB value $c(\cvec{x}) \in [0,1]^3$, semantic feature $f(\cvec{\cvec{x}}) \in \R^{C_{\mathcal{F}}}$ with feature dimension $C_{\mathcal{F}}$, TSDF value $s(\cvec{x}) \in [-1, 1]$, and corresponding log variance $u_{c}$, $u_{f}$, and $u_{s} \in \R$ to express uncertainty. By conditioning decoding networks on the unified, view-independent features~$\Phi$, the feature field captures scene priors, effectively enabling any-scene generalization.

We utilize differentiable volume rendering~\cite{mildenhall2020nerf} to project the predicted properties from 3D space into 2D for training. To apply volume rendering, we model the density at a point~$\sigma(\cvec{x})$ as the transformed {TSDF} following volume rendering methods for geometric reconstruction \cite{wang2021neus, azinovic2022neural, guo2022manhattan, yariv2021volumerendering}. Specifically, we adopt the Laplace cumulative distribution function from~\cite{yariv2021volumerendering} to define density as
\begin{equation*}
    \sigma_{\beta}(\cvec{x}) =
    \begin{cases}
        \frac{1}{\beta} (1 -  \frac{1}{2} \exp \left( \frac{{s}(\cvec{x})}{\beta} \right) ) & \text{if } {s}(\cvec{x}) < 0, \\
        \frac{1}{2 \beta} (\exp \left( - \frac{{s}(\cvec{x})}{\beta} \right) ) & \text{if } {s}(\cvec{x}) \geq 0, \\
    \end{cases}
\end{equation*}
where $\beta$ is a learnable parameter.
%Specifically, we render color, semantic features, and corresponding visual, semantic and spatial uncertainty using $c(\cvec{x})$, $f(\cvec{r})$, $u_c(\cvec{r})$, $u_f(\cvec{r})$, $u_s(\cvec{r})$, respectively. (\textbf{EXPLAIN})
For a ray $\cvec{r}(t)=\cvec{o} + t \cvec{d}$ with origin $\cvec{o}$ and view direction $\cvec{d}$, we render scene properties
\begin{equation*}
    \begin{split}
        \hat{C}(\cvec{r}) &= \int_{t_{\mathrm{n}}}^{t_{\mathrm{f}}} T(t) \; \sigma_{\beta}(\cvec{r}(t)) \; {c}(\cvec{r}(t)) \; \mathrm{d} t, \\
        \hat{F}(\cvec{r}) &= \int_{t_{\mathrm{n}}}^{t_{\mathrm{f}}} T(t) \; \sigma_{\beta}(\cvec{r}(t)) \; {f}(\cvec{r}(t)) \; \mathrm{d} t, \\
        \hat{D}(\cvec{r}) &= \int_{t_{\mathrm{n}}}^{t_{\mathrm{f}}} T(t) \; \sigma_{\beta}(\cvec{r}(t)) \; t \; \mathrm{d} t, \\
        \hat{U}(\cvec{r}) &= \int_{t_{\mathrm{n}}}^{t_{\mathrm{f}}} T(t) \; \sigma_{\beta}(\cvec{r}(t)) \; u(\cvec{r}(t)) \; \mathrm{d} t, \\
    \end{split}
\end{equation*}
\begin{equation*}
    \text{with } \quad T(t) = \exp \left( - \int_{t_{\mathrm{n}}}^{t_{\mathrm{f}}} \sigma_{\beta}(s) \mathrm{d} s \right),
\end{equation*}
where $C(\cvec{r}) \in [0, 1]^3$ is rendered RGB color, $F(\cvec{r}) \in \R^{C_{\mathcal{F}}}$ is a semantic feature with feature dimension $C_{\mathcal{F}}$, $D(\cvec{r}) \in \R$ is rendered depth, and $U(\cvec{r})$ is rendered log-variance of either color, semantic feature, or TSDF value.
Transmittance $T(t)$ quantifies the accumulated density up to $t$, and $t_{\mathrm{n}}$, $t_{\mathrm{f}}$ are the minimum and maximum bounding distances.

% \subsection{Feature field rendering}
% We utilize differentiable volume rendering to enable projections of properties into 2D image space. The density function $\sigma(\cvec{x})$ can be acquired by transforming the TSDF using the Laplace cumulative distribution function \cite{yariv2021volumerendering} following \cite{wang2021neus, azinovic2022neural, guo2022manhattan, yariv2021volumerendering}.
% We render different scene properties $\hat{P}(\cvec{r})$ (\eg color, CLIP features, ) of a ray $\cvec{r}(t)=\cvec{o} + t \cvec{d}$ with origin $\cvec{o}$ and view direction $\cvec{d}$ by
% \begin{equation}
%     \begin{split}
%         \hat{P}(\cvec{r}) &= \int_{t_{\mathrm{n}}}^{t_{\mathrm{f}}} T(t) \; \sigma_{TSDF}(\cvec{r}(t)) \; c(\cvec{r}(t)) \; \mathrm{d} t, \\
%     \end{split}
% \end{equation}

\subsection{Uncertainty-aware Supervision}

To supervise the visual and semantic properties of UniFField, we use ground-truth target RGB frames and pseudo-ground-truth semantic features. This pre-training task of novel-view reconstruction and feature prediction thus facilitates the learning of visual and semantic priors over any scene~\cite{fu2023scenepriors, ye2023featurenerf, qiu-hu-song-2024-geff}. For semantic feature supervision, we leverage knowledge distillation using MaskCLIP~\cite{zhou2022maskclip} as the teacher model~$\mathcal{F}$. Nevertheless, our model is designed to support any teacher model. While CLIP~\cite{radford2021learning} extracts image-level features, MaskCLIP allows extracting dense, patch-level features from CLIP, suitable for dense supervision. Aligning our unified features with those of CLIP allows for language-based querying in 3D at inference time. For geometric supervision, we apply TSDF learning~\cite{murez2020atlas, stier2023finerecon} by minimizing the difference between predicted and target TSDF values in 3D.

We supervise the model's color, semantic feature, and TSDF predictions by replacing the common loss function (e.g., L1 or L2 loss) with an uncertainty-aware loss function $\mathcal{L}^U$, which enables the learning of the uncertainty estimate alongside the model's output. We assume a Gaussian distribution of the model's output and utilize a heteroscedastic loss~\cite{kendall2017what}, typically used to quantify aleatoric uncertainty~\cite{pan2022activenerf} given by
\begin{equation} \label{eq:heteroscedastic_l2_loss}
    \mathcal{L}^{U}(y, \hat{y}, u) = \frac{1}{2} \exp(-u) \cdot \mathcal{L}(y, \hat{y}) + \frac{1}{2} u,  \\
\end{equation}
where $u$ is the predicted log-variance, $\hat{y}$ is the predicted mean and $y$ is the ground truth for an input $x$.
To control the trade-off between the prediction accuracy and the accuracy of predicted log-variance, we introduce a masked loss that blends between the heteroscedastic loss and the standard loss, given by
\begin{equation} \label{eq:masked_loss}
    \mathcal{L}^{\tilde{U}} = \sum_{i=1}^{M} \Bigl( m_i \cdot \mathcal{L}^{U} \bigl( y, \hat{y}, u \bigr) + (1-m_i) \cdot \mathcal{L} \bigl( y, \hat{y} \bigr) \Bigr),
\end{equation}
where $M$ is the number of samples $m_i \sim \mathrm{Bernoulli}(p)$, drawn from a Bernoulli distribution with probability $p$, that are used for supervision. %With this loss, our network learns to predict a combination of aleatoric and epistemic uncertainty.
% \begin{equation*}
%     \text{with } \quad m_i \sim \mathrm{Bernoulli}(p),
% \end{equation*}
% where $M$ is the number of samples used for supervision and $\mathrm{Bernoulli}(p)$ is the Bernoulli distribution with parameter $p$.
%Inspired by generalizable NeRF, we utilize novel view synthesis as a pre-training task to supervise the feature field with a masked L2 loss $\mathcal{L}^{\tilde{U}}_{\mathrm{rgb}}$ and facilitate the learning of visual and semantic priors \cite{fu2023scenepriors, ye2023featurenerf, qiu2024learning}.
% Moreover, we leverage knowledge distillation using MaskCLIP \cite{zhou2022maskclip} as the teacher model. 
%We therefore calculate the masked L2 loss $\mathcal{L}^{\tilde{U}}_{\mathrm{feat}}$ between the features of our model and the features of MaskCLIP, effectively aligning our unified features with those of CLIP \cite{radford2021learning}.
% We therefore minimize the difference between the features of our model and the features of MaskCLIP, effectively aligning our unified features with those of CLIP \cite{radford2021learning}.
%For geometric supervision, we apply TSDF learning \cite{murez2020atlas, stier2023finerecon} to perform geometric 3D reconstruction using the masked L1 loss between $\mathcal{L}^{\tilde{U}}_{\mathrm{tsdf}}$ between predicted and target TSDF values.

%We jointly supervise the model during pre-training on a large-scale dataset of scenes using the weighted sum of all three losses.
We train the model using RGB-D frame sequences from the ScanNet dataset~\cite{dai2017scannet}. During training, we first construct the feature field of a given scene using $M_{\mathrm{ref}}$ randomly sampled reference frames from the entire sequence. For supervision, we sample $M_{\mathrm{tgt}}$ additional target frames. For every target frame, we sample $N_{\mathrm{ray}}$ pixels to construct rays and use them for both the color loss $\mathcal{L}^{\tilde{U}}_{\mathrm{rgb}}$ and the semantic feature loss $\mathcal{L}^{\tilde{U}}_{\mathrm{feat}}$. For the {TSDF} loss $\mathcal{L}^{\tilde{U}}_{\mathrm{tsdf}}$, we use the resulting stratified point samples along the rays and supervise the {TSDF} values at these positions.% Therefore, the number of {TSDF} samples per frame is $N_{\mathrm{tsdf}} = N_{\mathrm{ray}} \cdot N_{\mathrm{bin}}$. In one training step, we compute the loss $\mathcal{L}^{\tilde{U}}$ for every target frame and average it, before passing it to the optimizer.

At inference time, we directly build the feature field of a novel scene and predict properties and uncertainty estimates in both 2D and 3D in a single forward pass. Incremental updates are made via a running average of the existing and new unified feature volumes from new RGB-D frames.%The outputs can then be directly used as input to downstream applications.
\begin{figure}[t!]
    \centering
    % --- First Figure ---
    \includegraphics[width=0.478\textwidth]{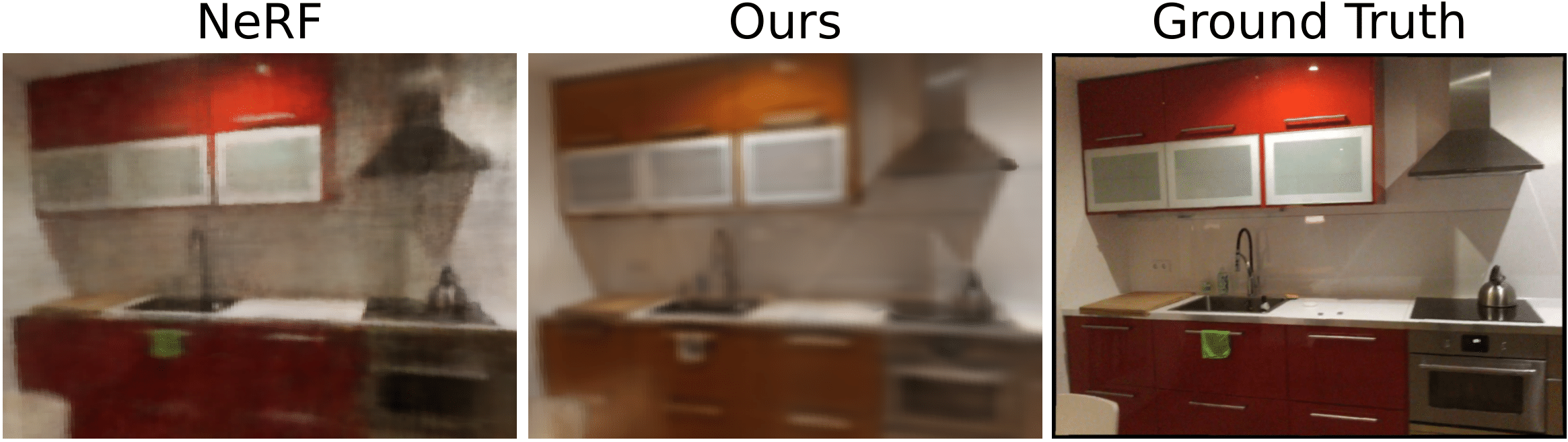}
    \caption[Novel view synthesis]{\textbf{Novel view synthesis}. 
    Here, NeRF is trained on 1658 reference frames, while our approach merges the feature volumes from reference frames without any optimization.
    }
    \label{fig:rgb}
    % --- Second Figure ---
    \vspace{0.5em}
    \includegraphics[width=0.478\textwidth]{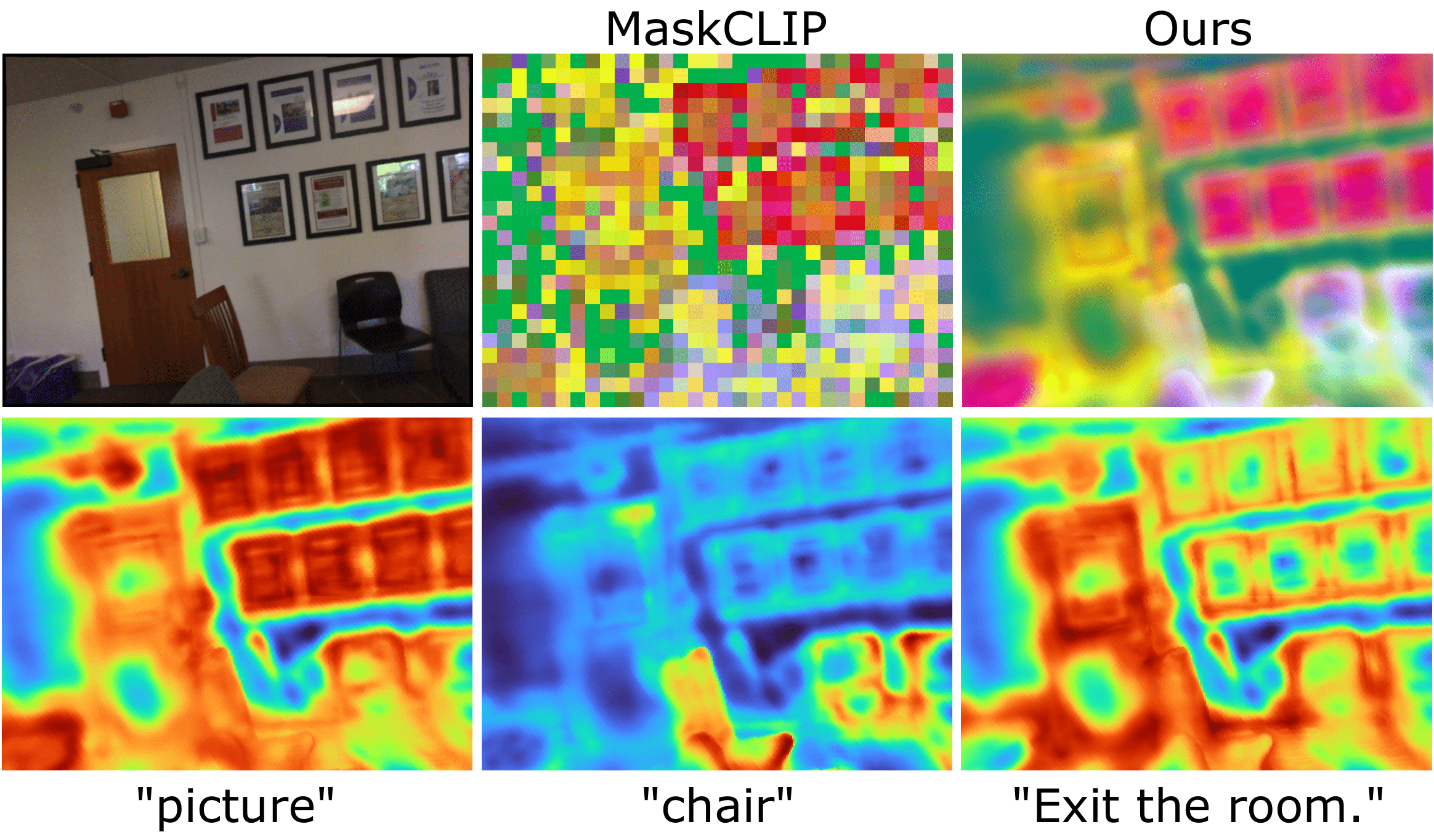}
    \caption[Semantic similarity search]{\textbf{Semantic similarity search}.
    %We show cosine similarity (red) between language queries and CLIP~\cite{radford2021learning} features predicted with our UniFField model in an unseen scene.
    We show CLIP~\cite{radford2021learning} feature maps predicted with MaskCLIP~\cite{zhou2022maskclip} and our UniFField model. The cosine similarity (red) between language queries and our predicted CLIP features is shown beneath.
    }
    \label{fig:text_sim_door_v1}
    % --- Third Figure ---
    \vspace{0.5em}
    \includegraphics[width=0.478\textwidth]{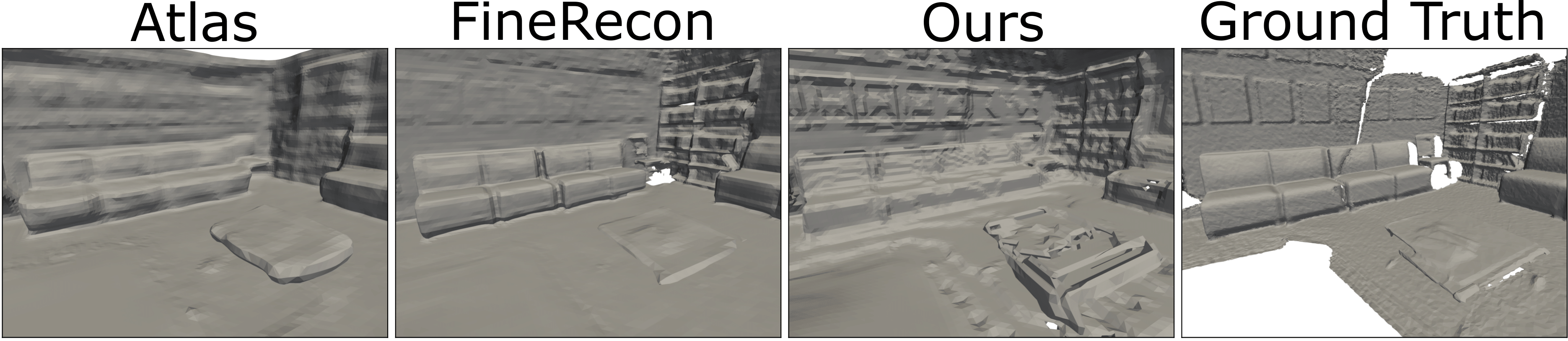}
    \caption[3D geometry alignment]{\textbf{3D geometric reconstruction}. Our UniFField model aligns with volumetric-based geometric reconstruction methods Atlas~\cite{murez2020atlas}, FineRecon~\cite{stier2023finerecon}, and produces complete geometry. UniFField utilizes depth guidance similar to FineRecon and captures finer but less smooth details than Atlas.}
    \label{fig:mesh}
    \vspace{-1.5em}
\end{figure}

\section{EXPERIMENTS}
% \textbf{Notes:}
% \begin{itemize}
%     \item Basic alignment with semantic, geometric, and visual properties (qualitative or quantitative)
%     \item Alignment of uncertainty with prediction error (Proof that the uncertainty is actually uncertainty - well calibrated) - semantic, geometric, visual uncertainties. Here we also compare with dropout ensemble. Ranks and Sparsification.
%     \item Performance on task: Object search (Active Perception / Uncertainty-guided exploration)
%     \item ALGO for active perception
%     \item Robot demo
%     \item Comparison with other baselines - naive baseline uses farthest point sampling, but ours uses a good combination of semantic and geometric uncertainties
%     \item Real robot metrics?: Trials, number of objects, number of frames used to build the representation
% \end{itemize}

We evaluate UniFField with the following experiments:% perform the following experiments to validate the effectiveness of UniFField:
\begin{itemize}
    \item First, we perform scene understanding experiments on unseen frame sequences from the ScanNet dataset to measure our representations' alignment with ground-truth visual, semantic, and geometric properties.
    \item Second, we evaluate our predicted uncertainties. We evaluate how well the uncertainty measure predicted by UniFField describes the prediction errors of the model.
    \item Third, we validate the ability of our representation to be used for active object search tasks in both simulation and on a real mobile manipulator robot.
\end{itemize}

%\subsection{Training Setup}
We train UniFField on ScanNet scenes~\cite{dai2017scannet}. % aiming towards generalization to indoor environments. % for 80 epochs. The training takes 18 hours on 4 NVIDIA A100 Tensor Core GPUs (80 GB). The feature grid has a resolution of $V_{\mathrm{x}} \times V_{\mathrm{y}} \times V_{\mathrm{z}} = 240 \times 240 \times 96$ and a voxel size of $v=0.04$ meters, covering a total volume size of $9.6 \times 9.6 \times 3.84$ meters. At test time, a voxel resolution of $400 \times 400 \times 104$ is used, corresponding to a volume size of $16 \times 16 \times 4.16$ meters.
For evaluation, we use a stream of input RGB-D reference frames of arbitrary length (e.g., a few frames to a few hundred ScanNet frames). Evaluations are performed on unseen scenes without per-scene optimization.
Creating a feature volume takes 0.04s per frame while extracting the TSDF and rendering a 640x480 feature map takes 1.26s and 7.70s respectively.

\subsection{Alignment with scene properties}

% TODO: if we omit Depth-guidance ablations -> check that it is not referenced in the text or any caption

We first verify that our unified feature field can be used as a general-purpose, task-agnostic scene representation for various 3D scene understanding tasks. We provide quantitative and qualitative results to verify alignment of UniFField with the ground-truth RGB, pseudo-ground-truth semantic features, and ground-truth TSDF of unseen ScanNet scenes.% demonstrate the capabilities and versatility of this approach by providing qualitative results on unseen scenes, exemplarily on three tasks.

%\paragraph{Novel View Synthesis}
%For the task of novel view synthesis, our method demonstrates the ability to generalize to unseen scenes and successfully recovers scene appearance without relying on per-scene optimization, as shown in \Cref{fig:rgb}.
% new:
For visual alignment, we compare against a NeRF~\cite{mildenhall2020nerf} as a reference point for a neural representation trained on a target scene. UniFField successfully recovers the scene's appearance without any optimization, as shown in \Cref{fig:rgb}. Furthermore, the quantitative results for a varying number of reference frames $M_{\mathrm{ref}}$ for the scene are shown in \Cref{tab:metrics_rgb}, demonstrating the effectiveness of UniFField in sparse data conditions with and without depth guidance (DG).% The results indicate that our method is more robust against sparse data conditions and does not rely on depth input.
\begin{table}[t]
    \centering
    \caption{Quantitative evals: alignment with scene properties 
    % (a) Visual alignment: novel view synthesis metrics. 
    % (b) 3D geometry alignment: reconstruction metrics.
    }
    \label{tab:metrics}

    % --- First subtable ---
    \begin{subtable}{0.95\linewidth}
        \centering
        \caption{Visual alignment: novel view synthesis metrics}
        \begin{tabular}{clccc} \toprule
            {$M_{\mathrm{ref}}$} &  {Method} & {PSNR $\uparrow$} & {SSIM $\uparrow$} & {LPIPS $\downarrow$} \\ \midrule
            \multirow{3}{*}{1658} & NeRF \cite{mildenhall2020nerf} &  \textbf{23.302}  &  \textbf{0.786}  &  \textbf{0.531} \\
                                  & Ours (w/o DG) & 18.602 & 0.752 & 0.575 \\
                                  & Ours & 18.216 & 0.752 & 0.569 \\ \midrule
            \multirow{3}{*}{50}   & NeRF \cite{mildenhall2020nerf} & 14.634 & 0.626 & 0.642 \\
                                  & Ours (w/o DG) & 16.060 & 0.701 & \textbf{0.632} \\
                                  & Ours & \textbf{16.259} & \textbf{0.705} & 0.639 \\ \midrule
            \multirow{3}{*}{25}   & NeRF \cite{mildenhall2020nerf} & 13.790 & 0.601 & 0.654 \\
                                  & Ours (w/o DG) & 14.881 & \textbf{0.679} & \textbf{0.633} \\
                                  & Ours & \textbf{15.013} & 0.671 & 0.648 \\ \bottomrule
        \end{tabular}
        \label{tab:metrics_rgb}
    \end{subtable}

    \vspace{0.65em}

    % --- Second subtable ---
    \begin{subtable}{0.95\linewidth}
        \centering
        \caption{Semantic alignment: feature alignment w/ MaskCLIP~\cite{zhou2022maskclip}}
        \begin{tabular}{cccc} \toprule
            {{CosineDist} $\downarrow$} & {$\mathrm{{MAE}}$  $\downarrow$} & {$\mathrm{{MSE}}$  $\downarrow$ } & {$\mathrm{{RMSE}}$ $\downarrow$}  \\ \midrule
            0.325  &  0.021  &  0.001  &  0.029  \\ \bottomrule
        \end{tabular}
        \label{tab:metrics_sem}
    \end{subtable}

    \vspace{0.65em}

    % --- Third subtable ---
    \begin{subtable}{0.95\linewidth}
        \centering
        \caption{3D geometry alignment: reconstruction metrics}
        \scalebox{0.815}{
        \begin{tabular}{lcccccc} \toprule
            Method & {Acc $\downarrow$} & {Comp $\downarrow$} & {Cham $\downarrow$} & {Prec $\uparrow$} & {Recall $\uparrow$} & {F-score $\uparrow$} \\ \midrule
            Atlas \cite{murez2020atlas}         & 0.128 & 0.110 & 0.119 & 0.647 & 0.382 & 0.476 \\
            Ours (w/o DG)                       & 0.612 & 0.146 & 0.379 & 0.483 & 0.220 & 0.299 \\
            FineRecon \cite{stier2023finerecon} & \textbf{0.111} & \textbf{0.037} & \textbf{0.074} & \textbf{0.901} & \textbf{0.428} & \textbf{0.578} \\
            Ours                                & 0.162 & 0.051 & 0.106 & 0.741 & 0.403 & 0.519 \\ \bottomrule
        \end{tabular}
        }
        \label{tab:metrics_geo_3d}
    \end{subtable}
    \vspace{-0.8em}
\end{table}

%\paragraph{Similarity Search}
%Previously: At inference time, UniFField can generate CLIP feature maps in unseen scenes that are spatially consistent and can be rendered at any resolution.
At inference, UniFField generates spatially consistent CLIP feature maps in unseen scenes that can be rendered at any resolution.
%Previously: They are sufficiently expressive to support semantic similarity search using cosine similarity with language queries, as shown in \Cref{fig:text_sim_door_v1}.
They are expressive enough to support semantic similarity search using cosine similarity with language queries, as shown in \Cref{fig:text_sim_door_v1}.
Quantitatively, over all 100 scenes in the ScanNet test set, the mean average error~(MAE) and squared errors~(MSE and RMSE) between normalized MaskCLIP and UniFField features are small~(\Cref{tab:metrics_sem}).

%\paragraph{Geometric Reconstruction}
For geometric alignment, we show geometric reconstruction results~(\Cref{tab:metrics_geo_3d}) following the evaluation protocol in~\cite{murez2020atlas} and compare with volumetric reconstruction methods Atlas~\cite{murez2020atlas} and FineRecon~\cite{stier2023finerecon} on all 100 scenes in the ScanNet test set. FineRecon only performs geometric reconstruction, and its performance serves as an upper-bound reference for UniFField since it uses a similar geometric architecture with depth guidance. The learned geometric priors of UniFField are similarly effective in producing complete geometry even in scene parts that were not observed~(\Cref{fig:mesh}). Our method competes with Atlas~\cite{murez2020atlas}, specifically, achieving better results in most metrics, including Chamfer distance and F-score. Although Atlas generally oversmoothes surfaces, our approach resolves higher detail at the cost of noisier geometry. % However, FineRecon outperforms both Atlas and our method in all 3D geometry metrics. It is worth noting that FineRecon and our method utilize depth in addition to RGB input to reconstruct geometry, whereas Atlas relies solely on RGB input.
The ablation of our approach without depth guidance (DG) reveals that reconstruction performance significantly depends on the additional depth input.%, resulting in overall similar quantitative and qualitative performance, despite relying on depth input.

\begin{table*}[ht]  % \textcolor{gray}{$0.0$}
    \centering
    %\captionsetup{justification=centering}
    \caption[Uncertainty evaluation]{\textbf{Uncertainty evaluation}. We compare our predicted uncertainties on the ScanNet dataset against dropout ensemble-based and random uncertainties of different modalities with the corresponding prediction error. For the correlation coefficient $\rho$, we additionally report the proportion of statistically significant correlation tests. %Predicted uncertainty generally outperforms dropout-based uncertainty across all error and uncertainty metrics, indicating a moderate correlation. The \gls{MAE} describes semantic uncertainty best.
    }
    \scalebox{1.}{
        \begin{tabular}{c|c|cc|ccc|ccc} \toprule
            \multirow{2}{*}{Space} & \multirow{2}{*}{Prediction Error} & \multicolumn{2}{c|}{\multirow{2}{*}{Uncertainty}} &   \multicolumn{3}{c|}{{AUSE} $\downarrow$} & \multicolumn{3}{c}{Correlation $\rho \uparrow$ (Significance $\uparrow$)} \\ \cline{5-7} \cline{8-10} %\hline
            & &  &  &  {{MAE}} & {{MSE}} & {{RMSE}} & {{MAE}} & {{MSE}} & {{RMSE}} \\ \hline
            \multirow{3}{*}{2D} & \multirow{3}{*}{Color} &\multirow{3}{*}{Visual} & Pred.    &    \textbf{0.213}  &  \textbf{0.243}  &  \textbf{0.233}  &  \textbf{0.474}  (0.97)  &  \textbf{0.481}  (0.97)  &  \textbf{0.481}  (0.97)  \\
                                  & &  & Drop.      &   0.263  &  0.310  &  0.289  &  0.375  (0.93)  &  0.380  (0.94)  &  0.380  (0.94)  \\
                                  & &     & Rand.       &   0.526  &  0.745  &  0.566  &  0.000  (0.04)  &  0.000  (0.04)  &  0.000  (0.04)   \\ \hline
                                       
            \multirow{3}{*}{2D} & \multirow{3}{*}{CLIP Feature} & \multirow{3}{*}{Semantic} & Pred.   &    \textbf{0.095}  &  \textbf{0.156}  &  \textbf{0.095}  &  \textbf{0.220}  (0.98)  &  \textbf{0.209}  (0.97)  &  \textbf{0.209}  (0.97)  \\
                                   & &  & Drop.    &   0.144  &  0.211  &  0.127  &  0.063  (0.54)  &  0.052  (0.54)  &  0.052  (0.54)  \\
                                   & &   & Rand.   &   0.170  &  0.238  &  0.141  &  0.000  (0.04)  &  0.001  (0.05)  &  0.001  (0.05)  \\ \hline

            % All voxels
            \multirow{3}{*}{3D} & \multirow{3}{*}{TSDF} & \multirow{3}{*}{Spatial}       & Pred.   &    \textbf{0.013}  &  \textbf{0.011}  &  \textbf{0.054}  &  0.561  (1.00)  &  0.561  (1.00)  &  0.561  (1.00)   \\
                               & &  & Drop.   &   0.164  &  0.184  &  0.326  &  \textbf{0.592}  (1.00)  &  \textbf{0.592}  (1.00)  &  \textbf{0.592}  (1.00)   \\
                               & &    & Rand.   &   0.965  &  0.967  &  0.969  &  0.000  (0.05)  &  0.000  (0.05)  &  0.000  (0.05)   \\ \bottomrule

        %    % Observed Voxels
        %    \multirow{2}{*}{TSDF} & \multirow{2}{*}{Spatial}   & Pred.   &    \textbf{0.045}  &  \textbf{0.048}  &  \textbf{0.097}  &  \textbf{0.841}  (1.00)  &  \textbf{0.841}  (1.00)  &  \textbf{0.841}  (1.00)   \\
        %                       &   & Drop.    &  0.159  &  0.150  &  0.266  &  0.817  (1.00)  &  0.817  (1.00)  &  0.817  (1.00)   \\ \bottomrule
        %                           %& Rand.   & Y &   0.856  &  0.901  &  0.854  &  0.000  (0.03)  &  0.000  (0.03)  &  0.000  (0.03)  \\  \bottomrule
    \end{tabular}
    }
    %\vspace{-1em}
    \label{tab:metrics_uncert_2d_rgb}
\end{table*}

\subsection{Uncertainty estimation}

\begin{figure}
    \centering
    \includegraphics[width=0.45\textwidth]{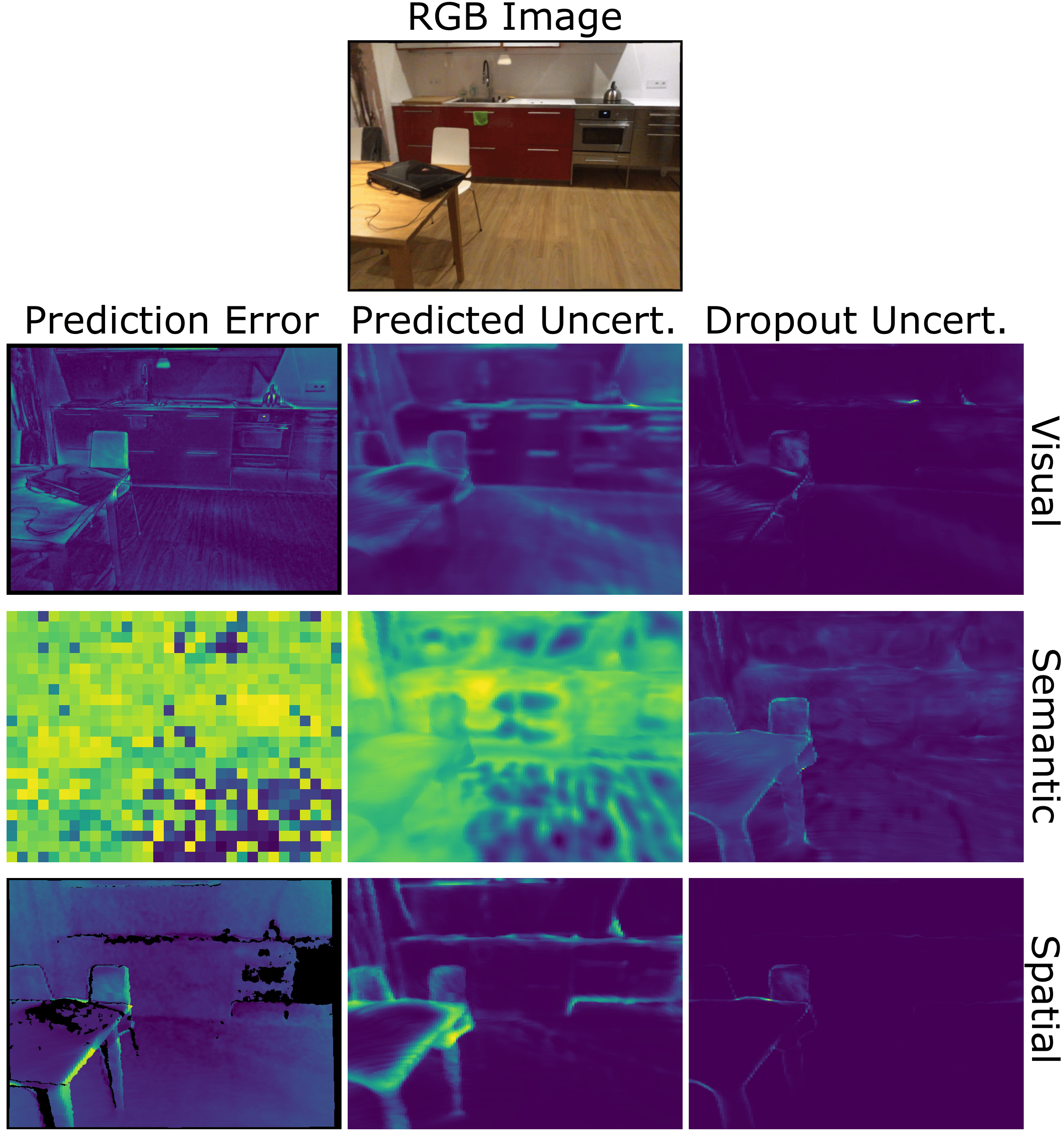}
    \caption[2D uncertainty]{\textbf{2D uncertainty}. 
    %We show the predicted and dropout-based uncertainty of different modalities and compare it against the prediction error.
    We compare different types and modalities of uncertainty against the prediction error.
    Visual uncertainty is most pronounced at the boundaries of objects, particularly in areas of high contrast differences. Semantic uncertainty is distributed across entire objects. Spatial uncertainty is most pronounced at object boundaries, where there is high depth contrast. The highest errors and uncertainties are colored yellow.}
    % \vspace{-0.5cm}
    \label{fig:uncert_2d}
\end{figure}

\begin{figure}
    \centering
    \includegraphics[width=0.476\textwidth]{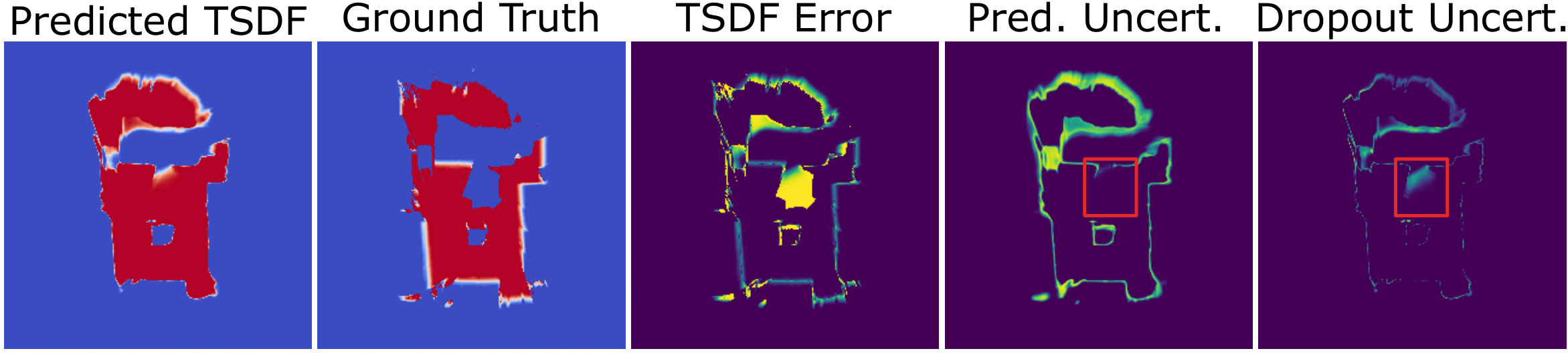}
    \caption[3D spatial uncertainty]{\textbf{3D spatial uncertainty}. We show slices of the voxel volumes at a constant height of $z=1.25$ meters. %predicted and target \TSDF, \TSDF error, as well as the predicted and dropout-based uncertainty volumes.
    Predicted uncertainty closely matches the TSDF error, while dropout-based uncertainty can detect errors caused by missing observations (red box). The highest errors and uncertainties are colored yellow.}
    \label{fig:uncert_3d_tsdf}
    \vspace{-1.0em}
\end{figure}

The key benefit of UniFField is the ability to model visual, spatial, and semantic uncertainties associated with the observed scene. To evaluate the quality of our learned uncertainties, we assess their alignment with the corresponding model prediction errors. Furthermore, we compare our learned uncertainties, which are a combination of aleatoric and epistemic uncertainty, against epistemic model uncertainties estimated using Monte Carlo dropout ensembles~\cite{kendall2017what, gal2016dropout}. To obtain dropout ensemble-based uncertainty, we add dropout operations to the 3D CNN convolutions and calculate the output variance over 10 forward passes. We evaluate over all 100 test scenes in ScanNet and, as in the previous subsection, perform visual and semantic evaluations for all frames in 2D by rendering our corresponding uncertainty outputs, while performing TSDF evaluations in 3D.
%For a well-calibrated uncertainty, the uncertainty estimates align with the prediction error.

We evaluate alignment with prediction errors---mean absolute error (MAE), mean squared error (MSE), and root mean squared error (RMSE)---using two metrics: (i) Area Under Sparsification Error (AUSE)~\cite{kondermann2008statistical,wannenwetsch2017probflow}, which is obtained by creating sparsification curves by progressively removing predictions with highest uncertainty and computing the error on the remaining predictions, (ii) Spearman’s rank correlation coefficient ($\rho$)~\cite{spearman1987proof}, which is a non-parametric measure that quantifies how well uncertainties track actual errors in a monotonic, rank-based manner. Unlike linear correlation measures, it does not assume linearity or a specific distribution. We consider a correlation statistically significant if the p-value is below a significance level $\alpha=0.05$. We also include a random uncertainty reference baseline, generated by sampling uniform uncertainty values to simulate a random ranking. It serves as a lower bound for the AUSE and correlation coefficient $\rho$ metrics.

% We analyze the alignment with prediction errors such as mean average error (MAE) and squared errors (MSE and RMSE) using two metrics: (i) the Area Under Sparsification Error~(AUSE)~\cite{kondermann2008statistical, wannenwetsch2017probflow} using sparsification curves.  to measure how well uncertainty identifies the prediction error. We use Spearman's rank correlation coefficient~$\rho$~\cite{spearman1987proof} to measure the strength and significance of the relationship between prediction error and uncertainty.

In \Cref{tab:metrics_uncert_2d_rgb}, the evaluation metrics are presented, which indicate a significant, monotonic relationship between the predicted uncertainties and their corresponding prediction errors. In most comparisons, the quantified uncertainties best describe the average deviations expressed with the MAE, compared to metrics that emphasize outliers (MSE or RMSE). A qualitative comparison reveals the behavior of different types of uncertainties across modalities, as shown in \Cref{fig:uncert_2d}.
%A qualitative comparison between MAE and uncertainty confirms this observation and reveals the behavior of different types of uncertainty across the modalities as shown in \Cref{fig:uncert_2d}.
%\Cref{tab:metrics_uncert_2d_rgb} shows the evaluation metrics for predicted uncertainty and dropout-based uncertainty across different error metrics.
Dropout ensemble-based uncertainty is less effective in identifying errors in the 2D domain, while it is slightly better for indicating 3D TSDF prediction errors across all voxels. Specifically, 3D spatial errors that arise due to unobserved areas can sometimes be better identified with dropout ensemble-based uncertainty in comparison to the predicted uncertainty, as illustrated in \Cref{fig:uncert_3d_tsdf}.

\subsection{Active object search with a mobile manipulator}

\begin{figure}
    \centering
    \includegraphics[width=0.47\textwidth]{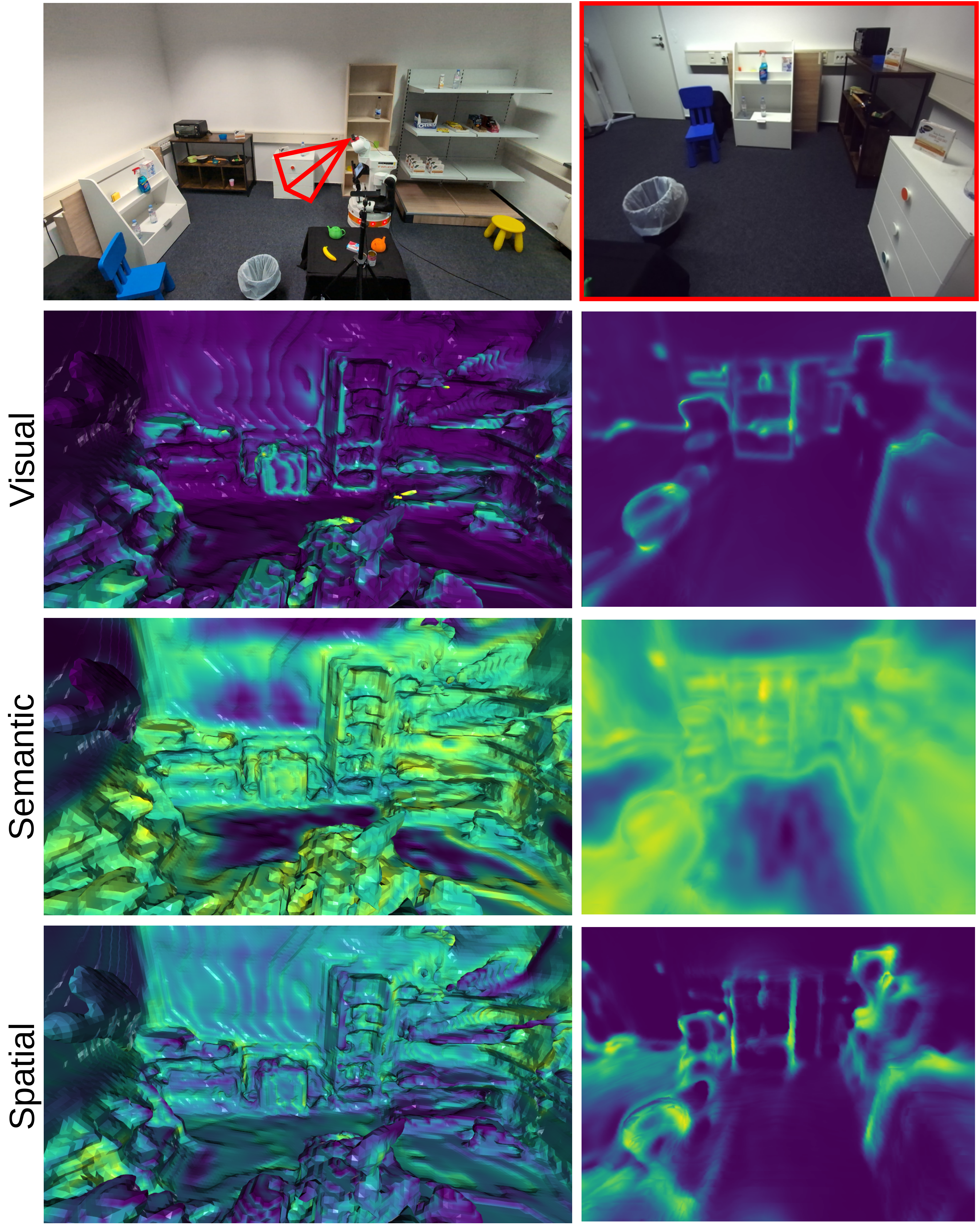}
    \caption[2D and 3D uncertainty]{\textbf{2D and 3D uncertainty}.
    Our model preserves spatial consistency in the predicted uncertainty and allows for 2D and 3D uncertainty estimation. The visualization is obtained by predicting uncertainties at 3D positions and mapping onto the nearest surface extracted from the predicted TSDF.
    }
    \label{fig:uncert_scene02}
    \vspace{-1.5em}
\end{figure}

We demonstrate a practical active object search task in the real world using a TIAGo mobile manipulator in an indoor environment. The robot is equipped with a head-mounted ZED2i RGB-D stereo camera. We analyze the captured uncertainties in the scene, highlight the flexibility of our approach in representing scene properties, and assess the robustness of our method in novel real-world data conditions.

The feature representation is created by collecting posed RGB-D observations using a robot object search policy in the indoor environment. We run inference to predict scene properties and use the semantic features to perform object search. Since our model is queryable at any 3D location, it allows predicting scene properties directly in 3D. As shown in \Cref{fig:uncert_scene02}, the properties of different uncertainties in 3D remain consistent with the rendered 2D uncertainty. We also observe low uncertainty across all modalities in simple-structured areas such as white walls or dark backgrounds. The drawer is similar to walls in terms of complexity of color and geometry, therefore exhibiting low spatial and visual uncertainty. However, it has a relatively higher semantic uncertainty, reflecting ambiguity, since it could also be interpreted as another piece of furniture.% Practically, the uncertainty can be used to guide exploration by focusing on areas of highest uncertainty.% Therefore, the predicted uncertainty volume can be restricted to the voxels near the TSDF zero-crossings and chosen by maximum without the need to extract a mesh.

\begin{figure} % TODO: (C: currently doing) show pick up of bottle; add multiple object search results or is one suffient?
    \centering
    \includegraphics[width=0.476\textwidth]{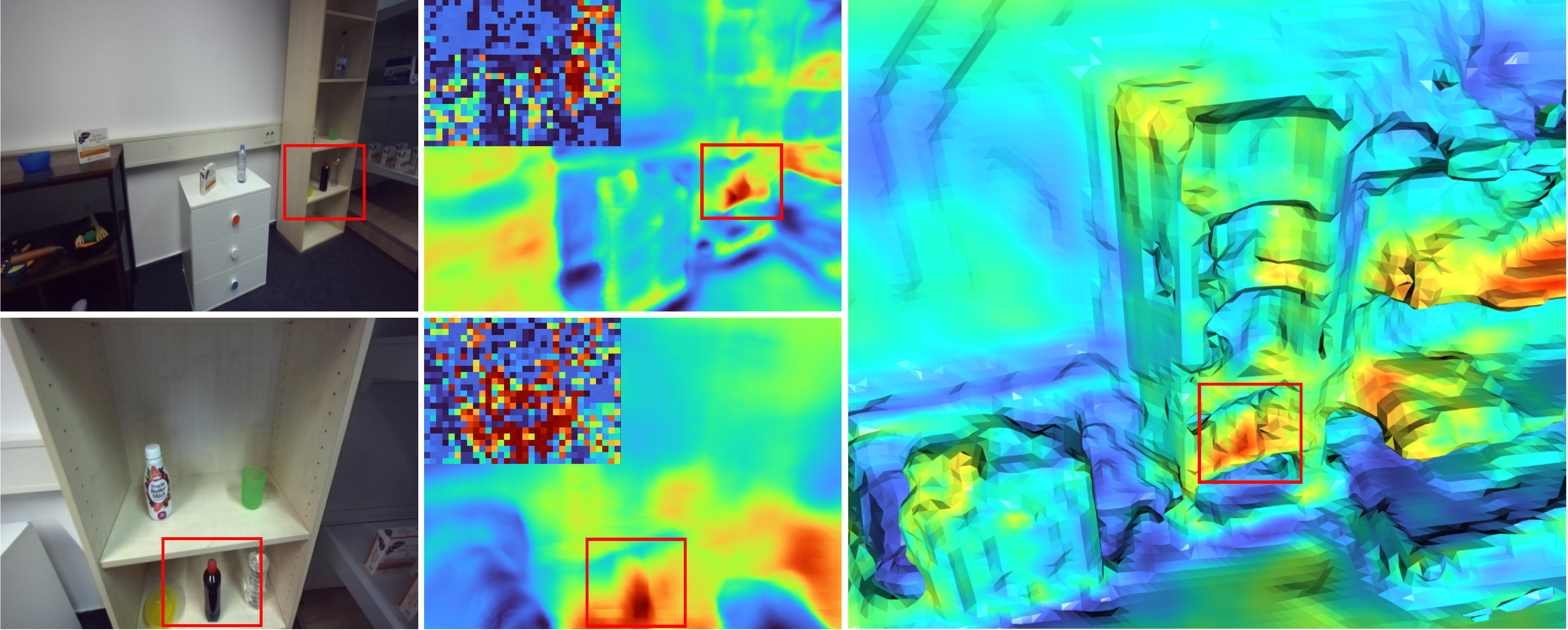}
    \caption[2D and 3D similarity]{\textbf{2D and 3D similarity}. The language similarity (red) for the query ``bottle on the shelf" is visualized in 2D and 3D space. We additionally show the similarity maps from the coarse feature map produced by MaskCLIP~\cite{zhou2022maskclip}. The model accurately localizes the queried object, demonstrating spatial consistency and high resolution. 
    }
    \label{fig:similarity_scene02}
    \vspace{-1.5em}
\end{figure}

To identify objects based on language queries, we first predict CLIP features for all voxels in our feature field and then calculate the cosine similarity between the features and the text encoding. 
To improve accuracy, we contrast the given positive text query with negative ones (e.g., ``wall" or ``ground") using a temperatured softmax, following~\cite{radford2021learning, qiu-hu-song-2024-geff}. 
%To improve accuracy, the features are additionally compared to the embeddings of negative text queries (e.g., ``wall" or ``ground"). We apply a temperatured softmax and sum up the probabilities of the positive query for the final similarity score, following~\cite{radford2021learning, qiu-hu-song-2024-geff}.
Similarly to 3D uncertainty, the similarity volume can be mapped onto the scene geometry. In \Cref{fig:similarity_scene02}, we illustrate the similarity search result. In contrast to MaskCLIP~\cite{zhou2022maskclip}, the predicted CLIP features are spatially consistent and do not depend on a particular good view of the scene.

\begin{figure}
    \centering
    \includegraphics[width=0.476\textwidth]{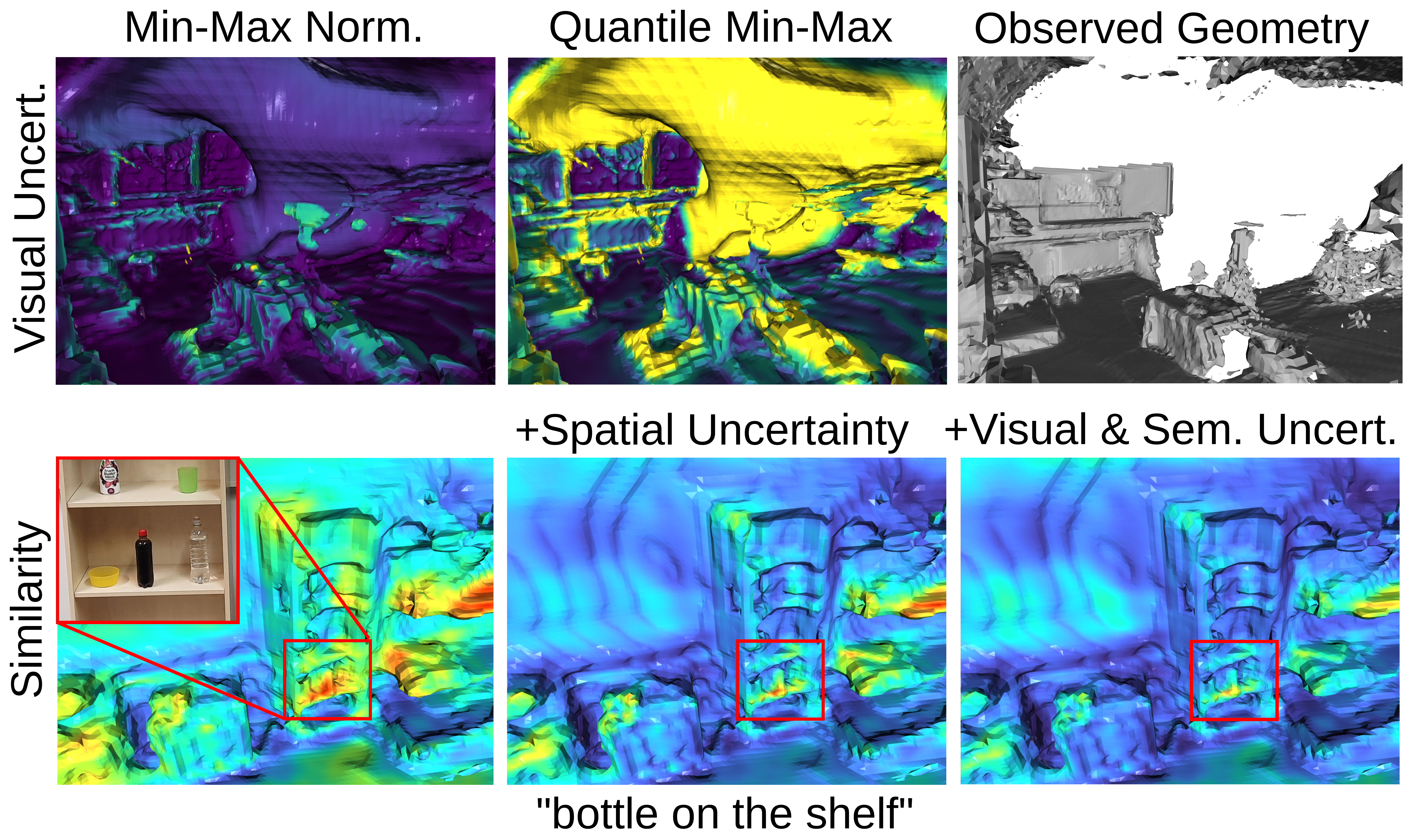} % sim_uncert_combinations_v2.pdf
    \caption[Uncertainty to improve exploration and similarity search]{\textbf{Uncertainty to improve exploration and similarity search}.
    We compare different methods of normalizing uncertainty across the scene and combining it with language similarity. 
    Quantile-based normalization restricts outliers, producing a measure that allows for the indication of unobserved scene geometry.
    Combining similarity using spatial uncertainty helps to improve the localization of a query object, while using all uncertainties lowers the overall similarity score for the specific target object.
    %We compare methods of combining language similarity with uncertainty. Weighting similarity by spatial certainty can help to improve the localization of a query object, while combining spatial, visual, and semantic uncertainties lowers the overall similarity score for the specific target object. After calculating every combination, min-max normalization is applied across the whole similarity volume.
    }
    \vspace{-1.5em}
    \label{fig:uncert_for_enhancements}
\end{figure}

% TODO: maybe as a new section/paragraph?
We design a rule-based robot policy that uses the scene information from UniFField in three phases. In an initialization phase, the robot collects a few observations from different viewing directions. Then, during an exploration phase, the scene areas of highest visual uncertainty are repeatedly localized. By sampling a location from surface regions with highest uncertainty, the next `look-at' position can be determined and approached, if not within a minimum distance. We find that replacing min-max normalization with quantile-based normalization when normalizing visual uncertainty can better indicate unobserved scene areas, as shown in \Cref{fig:uncert_for_enhancements}. After a fixed number of exploration steps, we transition to an exploitation phase. We localize the position of the most similar object according to the language query, while taking spatial uncertainty into account. In \Cref{fig:uncert_for_enhancements}, we show different methods for combining similarity and uncertainties. Since spatial uncertainty appears in areas of high depth contrast and complex geometry, we weight the similarity by the inverse of normalized spatial uncertainty. This puts less weight on geometrically uncertain regions for better localization of the target object. The shown combination of multiple uncertainties involves combining inverse uncertainties using a product and then using them as weighting for the similarity. A demonstration of the robot policy is available at \href{https://sites.google.com/view/uniffield}{https://sites.google.com/view/uniffield}.

%\section{CONCLUSION}
\section{CONCLUSION AND LIMITATIONS}

% Inference time (assuming a full table is too much):
% Building the feature representation requires 0.038 seconds per frame. Extracting the TSDF volume requires 1.259 seconds, and rendering a CLIP feature map of dimensions 640x480 requires from 7.696 seconds.
In this work, we introduced {UniFField}, a generalizable scene representation that quantifies uncertainty of different modalities from multi-view RGB-D data. Our experiments confirm that the representation generalizes to unseen scenes, enabling 3D scene understanding tasks while simultaneously allowing for uncertainty predictions that appropriately describe the corresponding prediction errors.

Nevertheless, the uncertainty estimates leave room for improvement, since we found that multiplicative combinations of uncertainty estimates do not always perform consistently, with effectiveness varying with language queries. Another limitation is that scaling up the model to enable larger-scale pretraining would hurt real-time performance. Our future work will thus focus on improving network inference speed and application to robotic tasks such as uncertainty-aware active object reconstruction.%, and will explore the suitability of VISUAL-FF for active perception.

% focus on different types and modalities of uncertainty 
% for usecases that benefit from more knowledge about the uncertainty in the environment and a flexible approach -> improvements for performance and quality possible

%                                   % the textheight of the last page by a suitable amount.
%                                   % This command does not take effect until the next page
%                                   % so it should come on the page before the last. Make
%                                   % sure that you do not shorten the textheight too much.

%%%%%%%%%%%%%%%%%%%%%%%%%%%%%%%%%%%%%%%%%%%%%%%%%%%%%%%%%%%%%%%%%%%%%%%%%%%%%%%%


% Appendices should appear before the acknowledgment.

% \section*{ACKNOWLEDGMENT}

% The authors gratefully acknowledge the scientific support and HPC resources provided by the Erlangen National High Performance Computing Center (NHR@FAU) of the Friedrich-Alexander-Universität Erlangen-Nürnberg (FAU) under the NHR project g101ea\_2. NHR funding is provided by federal and Bavarian state authorities. NHR@FAU hardware is partially funded by the German Research Foundation (DFG) – 440719683.

% BIBLIOGRAPHY
{
    \small
    \bibliographystyle{IEEEtran}
    \bibliography{refs}
}

\end{document}